# A Survey of Deep Fake Detection for Trial Courts

Naciye Celebi[1], Qingzhong Liu[1] and Muhammed Karatoprak[2]

[1]Department of Computer Science,
Sam Houston State University, Huntsville, TX, USA
[2]Department LLM in US Law, The University. Of Houston, Houston, TX, USA

## Abstract

*Recently, image manipulation has achieved rapid growth due to the advancement of sophisticated image editing tools. A recent surge of generated fake imagery and videos using neural networks is DeepFake. DeepFake algorithms can create fake images and videos that humans cannot distinguish from authentic ones. (GANs) have been extensively used for creating realistic images without accessing the original images. Therefore, it is become essential to detect fake videos to avoid spreading false information. This paper presents a survey of methods used to detect DeepFakes and datasets available for detecting DeepFakes in the literature to date. We present extensive discussions and research trends related to DeepFake technologies.*

## 1. Introduction

In the last two decades, image manipulation has achieved rapid growth due to the advancement of sophisticated image editing tools. The antiquated phrase "seeing is believing" is still the main perspective of validating such information. Hence, manipulated images convey misinformation and can be used to discredit people. Therefore, detecting manipulated images is increasingly essential and needs to be addressed.

A recent surge of generated fake imagery and videos using neural networks, so-called DeepFakes, are of great public concern and can now be easily created from scratch without leaving obvious perceptual traces. Open-source image editing tools lead to a large amount of generated Deepfake images and videos presented on social media, appearing a vital challenge for its' detection[1].

Recently, a new vein of fake image and video creation/generation methods known as DeepFake has been popular and attracted much attention. Many people can easily generate fake images and videos using smart phone and desktop applications such as FaceApp [2]. These applications have become much more prominent and advanced to generate extensive and realistic fake images, and videos can reveal the difficulty in determining their authenticity.Table1 shows the available DeepFake Generation tools. The step-by-step instructions and tutorials are easily available to create these fake images on the Internet. Therefore, these generation methods can be used for defamation, imitation, and misuse of facts. Furthermore, these DeepFakes can be widely and quickly disseminated on the Internet through social media. DeepFake takes as input an image or image of a specific person ('target') and outputs another image or video with the target's faces replaced with those of another person ('source'). DeepFake automatically maps the facial expressions of the source to the target. With decent post-processing, the output image has achieved a high level of realism.

Generative Adversarial Networks (GANs) have been extensively used for creating realistic images[3]. The adversarial framework of GANs can also be used in conditional scenarios for image swapping [4],[5],[6], which diversifies media synthesis. The goal of GANs is to generate similar-looking samples to those in the training set. More importantly, GANs generate these samples without access to the original images. As an example, we can complete an actor's movie who had recently passed away with the usage of GANs.

Table 1.  DeepFake Generation Tools

| Tools | Repositories |
| --- | --- |
| FaceApp | https://github.com/topics/faceapp |
| Faceswap-GAN | https://github.com/shaoanlu/faceswap-GAN |
| DFaker | https://github.com/dfaker/df |
| DeepFaceLab | https://github.com/iperov/DeepFaceLab |
| DeepFake-tf | https://github.com/StromWine/DeepFake-tf |

GANs are a combination of two neural networks (generator and discriminator) and interact only with the discriminative deep neural network to learn the data distribution. It competes to provide high-quality outputs similar to original inputs. GANs have been used to create new realistic images and videos and enhance those images. However, these machine learning algorithms, including GANs, can be misused to generate fake information to deceive people.

At the same time, however, the advancement of GANs has raised challenges to digital forensics. There is extensive concern about the impact of this technology when used maliciously. This issue has also received increasing public attention regarding disruptive outcomes to laws, politics, visual security, and society in general. Therefore, it is critical to look into practical visual forensics against threats from GANs.

Nowadays, social media has an active role in conducting an investigation. Many police officers can easily find a guilty person from a picture that can be found on his/her social media accounts. If a terrorist posts a video on YouTube expressing how he conducts a crime, investigators can easily find his identity. But how trustworthy are those founded images and videos? What if someone swapped a person's face with others on these images or videos? It is feasible that DeepFake challenges investigators and detecting these DeepFake images and videos are crucial.

## 2. RELATED WORKS

People are more and more interested in distinguishing GAN-generated media from real media. Many researchers have proposed various image forensics algorithms and tools to detect fake images, audio, and video. Many existing methods use the attributes of the image format and metadata information to determine the authenticity of the image. It is a challenging problem because attackers are also employing the latest image processing techniques to bypass wellknown forgery detection. Ding et al. [11] used deep transfer learning to extract a set of compact features and fed it into various classifiers such as SVM, random forest, and multi-layer perceptrons (MLP) for discriminating swapped face images from the genuine[11]. They are also different from existing methods that only provide detection accuracy. This study provides uncertainty for each prediction, which is critical for trust in the deployment of such faceswapping detection systems. They ranked their images according to their fakeness. McCloskey et al.[8] mentioned how much Image forensics is an increasingly relevant problem and they analyzed the structure of the generating network of a GAN implementation. They also showed that the network's treatment of color is markedly different from a real camera in two ways. They further showed that these two signs can be used to identify GAN-generated imagery from camera imagery, showing effective separation between GAN imagery and real camera images used[8] for training purposes. They used two different datasets produced in conjunction with the US National Institute of Standards and Technology's Media Forensics Challenge 2018[8]. These datasets addressed two different sets of GAN imagery: 1. GAN Crop images represent smaller image regions. 2. GAN Full images are mostly camera images, but some faces have been replaced by a GAN-generated face, similar to deep fakes. For both datasets, they computed the features over the entire image. Zhang et al. [12] generated swapped faces using identified faces in the LFW dataset [13]. They used Bag of Words and sped up robust features to create image features instead of using pixels. They tested different machine learning models on the created swap images such as Random Forests and SVM's. They achieved 93 percent accuracy but did not examine beyond their proprietary swapping techniques. Moreover, their dataset only has 5,000 real, 5,000 swapped images, which is relatively small compared to other works. Khodabakhsh et al. [14] examined the previously published methods. They created a new dataset, which is a combination of 53,000 images from 150 videos. The generated faces in their data set were collected using different techniques. They evaluated texture-based and CNN-based fake face detections. They also used smoothing and blending to make the generated faces more photo-realistic.

## 3. DEEP FAKE AUTHENTICITY

Authenticity, in Federal Courts, is governed by Rule 901(a) which provides "To satisfy the requirement of authenticating or identifying an item of evidence, the proponent must produce evidence sufficient to support a finding that the item is what the proponent claims it is, FRCP 34(b), which means all that is needed is evidence sufficient to convince a reasonable juror that a particular fact or event was more likely than not to have occurred. Before settling the authenticity of the evidence, the court must determine the relevancy of the evidence offered. Federal Rule of Evidence 401 provides that "Evidence is relevant if: (a) it has any tendency to make a fact more or less probable than it would be without the evidence; and (b) the fact is of consequence in determining the action." (1-1). After relevance of evidence offered was determined by courts, then the authenticity of it must be determined. The threshold for Rule 901 is not high "the court need not find that the evidence is necessarily what the proponent claims, but only that there is sufficient evidence that the jury ultimately might do so." Fed.R.Evid. 901(a). Federal Rule of Evidence 901(b) provides a list of 10 authentication methods satisfying the standard of proof for establishing authenticity, which is non-exhaustive and not limited to these methods. Some of the methods the list contains are testimony of a witness with knowledge, Fed.R.Evid. 401(a) and (b), opinion about a voice of United States v. Safavian, 435 F. Supp. 2d 36, 38 (D.D.C. 2006), evidence about a telephone conversation United States v. Vayner, 769 F.3d 125, 130 (2d Cir. 2014), and evidence about a process or system showing that it produces an accurate result. Evidence authentication methods listed under Rule 901(a) and (b) are also applicable to all kinds of digital evidence. Once the threshold requirement of authentication is satisfied, "the ultimate determination as to whether the evidence is, in fact, what its proponent claims is thereafter a matter for the jury. As an illustration, emails have been increasingly offered as evidence at trial. For authenticity purposes, a testimony of a witness(7a) could be used to determine if the email offered as evidence is authentic. The witness can testify that s/he saw the email in question created by the author or received by the person whom the proponent claims authored/received it. (7b)

## 4. INVESTIGATION OF DEEP FAKE

Due to advancements in technology, social media are currently used for additional evidence in court to establish support alibis and provide important information relevant to court cases. Social data gathering plays an increasingly valuable role in the evidence collection process.

Nowadays, evidence collected from social media could provide beneficial information for locating people in asset tracing investigations or establishing jurisdiction for legal proceedings.
A review of US case law shows limits on when the data that gathered from social media can be presented in a court of law as evidence. A murderer was captured on an image, and the image is introduced as evidence of the crime. The court directed that the evidence was allowed as long as the reliability of the digital evidence could be established. Image, audio, and video evidence could be presented under the silent witness theory that holds that the digital evidence(image, audio, or video) may be submitted as evidence without the requirement of a witness to verify its authenticity if it has been confirmed that the manner in which it was produced was reliable. The benefits of social media evidence are undeniable, ranging from primary and corroborative evidence to risk assessment and continuing evidence of activities. What about if the collected evidence is manipulated, so-called DeepFake? The challenge will arise for investigators. How their conducted investigation are accurate? In addition to reliability, courts held that social media evidence are admissible as long as their integrity, accuracy, and authenticity can be established. The question that follows is: how are these established?

Photographs have been manipulated and swapped images were popular and modified long before the digital image was invented. Investigators must compare the original photograph with the negative to verify the authenticity of an image. Digital image forensics has mostly focused on identifying low-level modifications in images, such as dropping or duplicating a frame or frames and regions swapping and copy-pasting a part of the primary image and placing them in other areas. Because of the existence of image, audio, and video manipulation, illegal convictions of perpetrators based on those social data will be at risk if an investigator does not consider the vulnerability of the social data manipulation. Investigators should be cognizant of how easily manipulated images, audio, and videos could present false evidence and lead to wrongful convictions. Some audio recordings present unique authentication issues, and their unaltered audio versions may or may not be available. Because social data that have been verified but are not what they imply could be wrongly admitted as evidence, investigators should utilize caution when introducing this type of evidence in the court of law. Corroborating evidence is required to show that the social data were not manipulated.

Historically, social data has been included to support eyewitness testimony. Currently, eyewitness testimony will be mandated to corroborate social data. Nevertheless, as machine learning and AI technology advances, the investigators may not be sufficient to authenticate evidence, because even expert witnesses may not be able to discern the manipulation made to social data.

Deepfakes are relative newcomers to digital media forensics. Images, audios, and videos must be authenticated before they are presented as evidence in a court of law; nevertheless, this method is difficult by the presence of Generative Adversarial Networks. Despite operating at their full potential, GANs are designed to improve their performance continually. As such, it is only a matter of moment before DeepFakes are so convincing that they are challenging to recognize as fakes.

## 4.1. Investigation Process

Digital forensics is essential for incident response strategy and provides an adequate response in a forensic manner [33]. These investigative steps are Examination, Identification, Collection, and Documentation. In [34], Tina et al. propose a new forensic model that allows the investigator to carry out a full forensic investigation by using the combination of cyber forensic and incident response models. The forensic process given in [34] consists of the following phases:

- Phase 1- Identification and Preparation: This is the initial phase of the proposed forensic process, and its purpose is to understand the social data which belong to the suspect.
- Phase 2- Identifying data sources: This phase is one of the most important phases of the process because it deals with identifying controllers of the system, the type of data that can be collected, and where the data can be collected. Data sources need to be identified when any type of data gathering is performed. Needless to say, documentation of the actions taken during this phase is critical and essential for a forensically sound investigation.
- Phase 3- Preservation, Prioritizing, and Collection: In this phase, the identified social data is collected from the known locations, and it is preserved and prioritized for the purpose of repeatability and presentation. In this phase, it is also critical to collect volatile data as it might be destroyed easily. For instance, data can be collected from the suspect's Facebook account, which may be deleted right after the investigation started.
- Phase 4- Examination: The purpose of this phase is the forensic examination of the collected evidence. In this phase, possible data filtering techniques can be used to reduce unrelated data. In this phase, the evidence data is simply surfaced using recovery techniques and tools for forensic analysis. Later on, the evidence's authenticity needs to be tested with the usage of the DeepFake Detection Model/Tool. If that evidence is not manipulated, then the evidence is ready for analysis.
- Phase 5- Analysis: This phase includes recovered forensic artifacts and collected evidential data in order to develop a timeline of the events/incidents. The actual analysis of the data is performed in this phase.
- Phase 6- Reporting and Presentation: This phase is the collection of findings during the examination and analysis phases. It should include the chain of custody documents to protect the admissibility and reliability of the evidence.
- Phase 7- Review Results In this phase, all the investigative process is reviewed for a comprehensive look to identify inculpatory or exculpatory data. The investigator may prove or disprove certain explanations made earlier.

## 4.2. Digital Evidence and Authentication

The rapid development of technology since the late 20th century and the use of technological devices such as phones, computers, and digital storage tools have made a significant transition in people's daily lives and changed their habits. Digital devices are everywhere in today's world, helping them in their everyday life, from communication to education, health, productivity, and so on. For example, small devices such as laptops, tablets, and smartphones took shelves full of encyclopedias' places in searching for information. The courtrooms have also been affected by these developments in collecting and handling evidence. The 2006 amendments to the Federal Rules of Civil Procedure (FRCP) included a new term of Electronically Stored Information (ESI) which refers to any documents or information stored in electronic form. Most common examples of ESI include word processing documents, digital photographs, videos, emails, text messages, and social media postings. After an amendment that included ESI into the Federal Rules of Civil Procedure, digital evidence has been offered more commonly at trial and has dramatically increased in volume as parallel to the increasing use of technological devices. Digital evidence includes but is not limited to emails, photographs, videos, texts, Facebook posts, info derived from websites, etc.

In terms of authenticity, accepting digital evidence also brought some challenges to courts since the authenticity of digital evidence is usually a central battleground to determine its admissibility. Authenticity in Federal Courts is governed by Rule 901(a) which indicates "to satisfy the requirement of authenticating or identifying an item of evidence, the proponent must produce evidence sufficient to support a finding that the item is what the proponent claims it is" which means all that needed is evidence sufficient to convince a reasonable juror that a particular fact or event was more likely than not to have occurred. Before ruling about the authenticity of the evidence, courts must determine the relevancy of the evidence brought before the court. Federal Rule of Evidence 401 provides that "Evidence is relevant if: (a) it has any tendency to make a fact more or less probable than it would be without the evidence, and (b) the fact is of consequence in determining the action.". After the relevancy of evidence was determined by courts, the authenticity of it must be concluded. The threshold for Rule 901 is not high which the court in the United States v. Safavian case held that "the court need not find that the evidence is necessarily what the proponent claims, but only that there is sufficient evidence that the jury ultimately might do so.". Once the threshold requirement of authentication is satisfied, "the ultimate determination as to whether the evidence is, in fact, what its proponent claims is thereafter a matter for the jury. Federal Rule of Evidence 901(b) provides a nonexhaustive list of authentication methods satisfying the standard of proof for establishing authenticity. Some of the methods the list contains are testimony of a witness with knowledge, opinion about a voice, evidence about a telephone conversation, and evidence about a process or system showing that it produces an accurate result[9]. There are no substantial changes to the Federal Rules of Evidence in terms of authenticating digital evidence and evidence authentication methods listed under Rule 901(b) also applies to all kinds of digital evidence. As an illustration, emails have been increasingly offered as evidence at trial. For authenticity purposes, a testimony of a witness[10] could be used to determine if the email offered as evidence is authentic. The witness can testify that s/he saw the email in question created by the author or received by the person whom the proponent claims authored/received it.

Rule 902 allows certain types of evidence to be self-authenticating which means no need for any extrinsic evidence for the purpose of authenticity, unlike witness testimony given as an example above. However, the opponent of the evidence may always challenge the authenticity. Unlike rule 901(b)'s non-exhaustive list of authentication methods, self-authentication methods listed under 902 are limited. In 2017, sections FRE 902 and 902 were added to the Federal Rules of Evidence as categories of self-authenticating digital records. Added FRE 90213 is related to Certified Records Generated by an Electronic Process or System and provides that "A record generated by an electronic process or system that produces an accurate result, as shown by a certification of a qualified person that complies with the certification requirements of Rule 902 (11) or (12). The proponent must also meet the notice requirements of 902(11).". Also FRE 902(14) regulates Certified Data Copied from an Electronic Device, Storage Medium, or File which provides "Data copied from an electronic device, storage medium or file if authenticated by a process of digital identification, as shown by a certification of a qualified person that complies with the certification requirements of Rule 902(11) or (12). The proponent also must meet the notice requirements of 902(11).". These newly added sections will allow both sides to authenticate certain types of Electronically Stored Information(ESI) without the need to offer any extrinsic authentication method provided in 901(b). For example, a party could introduce GPS data as evidence if s/he could introduce an authentication certificate pursuant to FRE 902(13), or a party could introduce text messages certifying these text messages are the same as the originals. With these amendments, no live witnesses are required to authenticate certain machine-generated data in some cases. However, the ability to fabricate digital data has made the authenticity of digital evidence a vital issue. With the ongoing advances in technology, manipulation of digital evidence has become easier for perpetrators. Now, it is possible as well as easy to create realistic-looking videos, so-called Deepfakes, that show people saying and doing things that they never said or did. A deepfake video utilizes machine learning to transform people's image, audio, and video so these data resemble someone else and can manipulate people's words and actions. The more video and audio footage of real people that can be fed into the generated model, the more convincing the result will be[35]. The advent of so-called "DeepFake" social data has recently been used widely in political disinformation campaigns to injure political opponents. In 2018, a deepfake video showing Barack Obama saying words he never said went viral on Youtube, also, in another video, the audio of Nancy Pelosi was altered by slowing down the audio to show she seems to be slurring. Most recently, a digitally altered, so-called deepfake, video of Queen Elizabeth delivering an alternative Christmas speech showing her dancing during the speech left too many people behind believing the video is real.

All these examples made people think if the antiquated phrase "seeing is believing" still be the main perspective to validate the information.

### 4.3. Deepfakes in Courtrooms

As an inevitable consequence of the widespread use of technology, it is not hard to anticipate that we will see more deepfakes in pre-trial and trial stages of courtrooms in near future. In a recent child custody case in the United Kingdom, the voice of a child's father was manipulated by the child's mother showing the father was heard making violent threats towards his wife. After expert examination, the recording was found as manipulated to include words not used by the father. The mother herself used a software program and online tutorials to put together a plausible audio file[36]. The case illustrates a good example of what kind of problems are laying ahead of courts. As the case illustrates, there is no need to be an expert to manipulate digital evidence because even a layperson is able to create DeepFakes by watching a few tutorials or utilizing ready-to-use programs. Deepfakes will have undesirable effects on courts that "The foreseeable effects will be both direct and indirect"[37] says Riana Pfefferkorn, professor of expertise in the area of cyber security. As a direct effect, DeepFakes can cause some additional caseload to courts by giving rise to new tort claims on the basis of this deepfake evidence brought to courts. In the above-mentioned child custody case, for example, the father may bring a tort claim against the mother for the deepfake evidence produced by her. As an indirect effect, the alleged deepfake will not be the main reason for the lawsuit but rather the deepfake will be just another piece of evidence in the course of litigation, where video evidence is already common[38]. Determining if the introduced video and other digital media evidence are either admissible or not will cause courts to spend too much time on this particular piece of evidence as well as will be costly when expert opinion is needed. Video is a widely used digital evidence in courts but the trustworthiness of videos should be questioned very often no matter which source the videos come from. Protecting the integrity of evidence, such as a video, is as important as getting the evidence itself, and in some cases, this integrity could be harmed by people who purposefully attack this evidence. A security firm consultant, Josh Mitchell, analyzed five body camera models from five different companies which all companies sell their devices to law enforcement groups around the US. The result of the study showed that four of the five body camera devices are remotely accessible and have security issues that allow an attacker to download footage off a camera, add modifications, or erase some footage from the video that the attackers don't want law enforcement to see, and then re-upload it, leaving no trace of the change[37]. Even though courts can authenticate these video evidence by one of the non-exhaustive methods listed under FRE 901(b) such as a witness with knowledge, not necessarily the person who took the video, providing when, where and under what circumstances the video was taken, bigger problems give rise under the jurisdictions that accept so-called silent witness theory which means photo or video speaks for itself. Silent witness theory focuses on the automatic operation of the video and allows photographic evidence to be admitted without verification of accuracy from eyewitnesses. Under jurisdictions that accept silent witness theory a photo or video, even it is a deepfake, may get into evidence more easily even confronted by the other side. Courts are already familiar with forgery and have rules of evidence to deal with the authenticity of evidence[38]. But, technology's widespread accessibility and the capacity to produce deepfakes by manipulating videos is growing at a meteoric rate compared to the ability as well as in terms of available tools used to distinguish deepfakes from genuine evidence. Soon it will be a big hurdle for courts to distinguish genuine evidence from deepfakes as the software programs and instructive materials to make deepfakes continue to improve at that rate. On the other side, there are available remedies to keep the negative effects of deepfakes at a minimum in courtrooms. For example, a total ban of deepfakes should be thought on but a total injunction of creating deepfakes most probably will face the First Amendment's freedom of speech clause and will be hard to prevail[39]. Still, some states enacted laws focusing on particular issues caused by deepfakes[21]. As a more effective remedy, more sophisticated tools that use Artificial Intelligence(AI) may be utilized by courts to detect manipulated evidence, so-called DeepFakes, from genuine evidence. In 2018, the Defense Department has produced the first tools for catching deepfakes[40]. Also, big tech companies, such as Microsoft, Facebook, Google, Amazon have been investing and developing AI-based detection methods[41].

Consequently, the importance of digital forensics and AI tools to detect deepfakes from genuine evidence has been becoming more important than ever. Courts need to use AI against AI to get better results in terms of authenticating evidence. Even though it is still not a big hurdle to distinguish authentic evidence from deepfakes, experts warn that deepfakes will be indistinguishable from real evidence at that pace. "In January 2019, deep fakes were buggy and flickery," said Hany Farid, a UC Berkeley professor and deepfake expert. "Nine months later, I've never seen anything like how fast they're going. This is the tip of the iceberg." "We are outgunned," said Farid. "The number of people working on the video-synthesis side, as opposed to the detector side, is 100 to 1."[42]

## 5. AVAILABLE DATASETS

Deepfakes are potential counter-productive to society, security, and privacy. Many researchers have proposed methods for detecting DeepFakes. Early attempts were based on handcrafted features collected from artifacts and deviations of the fake image combination process. On the other hand, recent methods mainly focused on automatic detection system, which consists of Deep Learning models. The detection methods are focused on a binary classification situation where classifiers are used to classify between real videos and tampered ones. Real and fake video and images dataset are required in order to apply the deep learning models to train classification models. The number of fake videos and images is available, but it is still limited in terms of setting a benchmark for validating various detection methods.

- Celeb-DF[15]: This dataset is comprised of 590 real videos and 5, 639 fake videos

- FaceForensics++[16]: his dataset includes a subset of DeepFakes videos, which has 1, 000 real YouTube videos and the same number of synthetic videos generated using faceswap.

- DeepFake Detection[17]: The Google/Jigsaw DeepFake detection dataset is a subset of 3, 068 DeepFake videos generated based on 363 original videos of 28 consented individuals of various genders, ages, and ethnic groups. The details of the generation algorithm are not disclosed.DeepFake Detection Preview: The preview of the Facebook DeepFake detection challenge dataset is a subset of the DeepFake detection challenge, which has 4, 113 DeepFake videos created based on 1, 131 original videos of 66 consented individuals of various genders, ages, and ethnic groups. This dataset is created using two different synthesis algorithms, but the details of the synthesis algorithm are not disclosed.

- HOHA Dataset[18]: This dataset is a combination of 300 real videos.

- UADFV[19]: This dataset contains 49 real YouTube and 49 fake videos. The DeepFake videos are generated using the FakeAPP.

- DF-TIMIT[20]: This dataset includes 640 fake videos generated with faceswap-GAN and is a subset of the Vid-TIMIT dataset. The videos are divided into DF-TIMIT-LQ and DF-TIMITHQ datasets, with synthesized faces of size $64 \times 64$ and $128 \times 128$ pixels, respectively.

- VTD Dataset[21]: This dataset includes 320 real videos collected from YouTube.

## 6. AVAILABLE DETECTION MODELS

Guera et al.[7] proposed a system that detects fake videos. They used the convolutional neural network (CNN) for extraction purposes and used these extracted futures to train a recurrent neural network (RNN) that learns to classify if a video has been subject to manipulation or not. They collected 300 deepfake videos from some video-hosting websites. Then, they randomly selected 300 videos from HOHA dataset. Their system had achieved more than 95 percent accuracy. They reached high accuracy because the trained dataset size is not big enough. Typically, CNN and RNN models require more datasets to train. [9] Korshunov et al. introduced a publicly available dataset for Deepfake and named as VidTIMIT database. In their paper, they introduce open-source software based on GANs to produce the Deepfakes, and they showed that training and combining parameters could significantly affect the quality of the produced videos. To demonstrate this result, they generated low and high-quality images and videos using separately tuned parameter sets. They pointed out that recently, accessible face identification systems are vulnerable to DeepFake images and videos, with 85.62 percent and 95.00 percent false acceptance rates. They also stated that it is necessary to identify DeepFake images and videos. They found out that the audio-visual method based on lipsync inconsistency detection could not identify Deepfake videos. Their experiments prove that GAN-generated DeepFake images and videos are challenging. They also mentioned that the further development of faceswapping technology would make it even more challenging. [10] Agarwal et al. described a forensic method that shows facial expressions and actions that symbolize an individual's speaking pattern. They also discussed how DeepFakes are designed and can, therefore, be used for authentication. They used the open-source facial expression analysis toolkit OpenFace2 to extract facial and head movements in a video. They also used t-SNE method for visualization of the 190-dimensional characteristics for Hillary Clinton, Barack Obama, Bernie Sanders, Donald Trump, Elizabeth Warren. They compare CNN and FaceForensics++ models. They detected that FaceForensics++ operates reasonably well on face-swapping but did not conclude lip-sync deep fakes.

Li et al. [22] proposed a new deep learning-based model that can efficiently identify DeepFakes. Their proposed method is focused on the new DeepFake algorithm that can only produce images and videos of limited resolutions, which require to be further distorted to meet the real faces in the source image and video. In this study, the authors pointed out that DeepFakes could be effectively obtained by convolutional neural networks (CNNs). They presented that their proposed model does not need DeepFake generated images as false positives since they target the artifacts in affine face-swapping as the distinctive feature to distinguish real and fake images[22]. They also stated their success in this study, (1) Such artifacts can be simulated instantly using simple image processing models on an image to make it a negative example. (2) Since such artifacts generally exist in DeepFake videos from various sources, their method is more robust than others.

Fernandes et al [23] mentioned that it is become essential to detect fake videos to avoid the spread of false information. In this paper, they used the heart rate of fake videos to distinguish between original and fake videos. They obtained the heart rate of original videos and trained the state-of-the-art Neural Ordinary Differential Equations (Neural-ODE) model. Then they created DeepFake videos using DeepFake generation tools. The average loss obtained for ten original videos is 0.010927, and ten donor videos are 0.010041[23]. The trained Neural-ODE was able to predict the heart rate of our 10 DeepFake videos generated using DeepFake generation tools and 320 DeepFake videos of the deepfakeTIMIT database. This is the first attempt to train a Neural-ODE on original videos to predict the heart rate of fake videos.

Amerini et al. [24] proposed a new forensic technique that can be discerned within fake and real image and video sequences. They compared state-of-the-art techniques that resort to every image frame, and their proposed selection of visual movement fields employs reasonable interframe variations. They additionally used the feature which is learned by CNN classifiers. Preliminary results were collected on the FaceForensics++ dataset and received quite promising performances. FaceForensics++ has been used with three automated image and video manipulation methods: Deepfakes, Face2Face, and FaceSwap. 720 of the image and videos are used for training purposes, 120 for validation, and another 120 for testing. They received VGG16: 81.61percent ResNet50: 75.46 percent on the Face2Face dataset.

In this paper, Afchar et al.[25] introduced a method to automatically and effectively distinguish DeepFakes, and they essentially concentrated on two contemporary methods used to generate manipulated images and videos: Face2Face and DeepFake. They stated that recent image forensics methods are normally not well suitable for image videos due to the concentration that completely discredits the data. Therefore, in this paper, the authors developed a deep learning approach and performed two networks, both with a low number of layers, to focus on images' mesoscopic properties. They also evaluated those fast networks on both Face2Face datasets. The experiments show a quite well detection rate with more than 98 percent for Deepfake and 95 percent for Face2Face.

Jeon et al.[26] proposed FakeTalkerDetect, which is based on siamese networks to detect the talking head with few-shot learning. Unlike conventional methods, they also proposed to use pre-trained models with only a few real image datasets for fine-tuning in siamese networks to efficiently detect the fake images in a highly unbalanced data setting. The FakeTalkerDetect achieves an overall accuracy of 98.81 percent accuracy in detecting DeepFakes generated from the latest neural talking head models. In particular, their preliminary work also demonstrates the effectiveness of the highly unbalanced dataset.

Xuan et al.[27] mentioned that GAN-generated images and videos are getting more practical with the high-quality generation, and it is infeasible for individual eyes to detect. On the other hand, researchers generate methods to identify these DeepFakes and guarantee social data's reliability. They also stated that the detection of DeepFake images and videos is a fundamental research problem. This paper explored this problem and proposed using preprocessed images and videos to embed a forensic CNN model. By performing image-level preprocessing to both real and fake images and video training sets, the CNN model is forced to learn to distinguish the fake and real images and videos.

Sabir et al.[28] extract the most suitable strategy for combining varieties in the RNN model and domain-specific face preprocessing methods through experimentation to obtain a state-of-the-art review on the FaceForensics++ dataset. The authors endeavored to detect Deepfake, Face2Face, and FaceSwap tampered faces in video streams in this study. Evaluation is performed on the recently introduced FaceForensics++ dataset, and they developed the previous state-of-the-art with up to 4.55 percent of accuracy.

Yang et al.[29] introduced a new method to detect DeepFakes. Their approach is mainly focused on the observations that Deep Fakes are created by splicing combined face area into the primary image/video. Hence, they suggested that errors may be exposed when 3D head postures are determined from the images and videos. They also conducted analyses to interpret the DeepFakes and, moreover, develop a classification method based on this suggestion. An SVM classifier is evaluated with the usage of real face images and videos.

Koopman et al. [30] declared that DeepFake poses forensic challenges concerning the authenticity of image and video evidence. As a result, photo response non-uniformity (PRNU) analysis is examined by authors for its effectiveness at identifying DeepFakes. The PRNU analysis shows an important variation in mean normalized cross-correlation scores between authentic image/videos and Deepfakes.

## 7. CONCLUSION

Recently, many people have begun to apprehend the existence of DeepFakes. They are aware that DeepFakes convey misinformation and can be used to misled people due to the dissemination of DeepFakes are increasingly approachable, and social media platforms can spread the DeepFakes quickly. Can awareness of DeepFakes helps people to validate such information, or it may distress, and adverse effects to targeted, heighten disinformation? The answer is really easy; it consumes the trust of people in media content, as seeing them is no longer believing in them. More often, DeepFakes do not need to be expanded to the massive audience to cause harmful effects.